\pdfoutput=1

\documentclass[11pt]{article}
\usepackage[inkscapelatex=false]{svg}

\usepackage[preprint]{acl}

\usepackage{times}
\usepackage{latexsym}

\usepackage[T1]{fontenc}

\usepackage[utf8]{inputenc}

\usepackage{microtype}

\usepackage{inconsolata}

\usepackage{graphicx}

\usepackage{graphicx}
\usepackage{dsfont}
\usepackage{amsmath}
\usepackage{amssymb}
\usepackage[english]{babel}
\usepackage{amsthm}
\usepackage{multirow}
\usepackage{makecell}
\usepackage{graphicx}
\usepackage{verbatim}
\usepackage{booktabs}
\usepackage{arydshln}
\usepackage{xcolor}
\usepackage{paralist}
\usepackage{colortbl}
\usepackage{cleveref}
\usepackage{transparent}
\usepackage{nicefrac}
\usepackage{subcaption}
\usepackage{algorithmic,algorithm}
\usepackage{enumitem}
\usepackage{tcolorbox}
\tcbuselibrary{breakable}
\usepackage{multicol}
\usepackage{CJKutf8}
\newcolumntype{C}[1]{>{\centering\let\newline\\\arraybackslash\hspace{0pt}}m{#1}}
\newcolumntype{R}[1]{>{\PreserveBackslash\raggedleft}p{#1}}
\newcolumntype{L}[1]{>{\PreserveBackslash\raggedright}p{#1}}

\definecolor{nred}{RGB}{196, 38, 11}
\definecolor{nblue}{RGB}{41, 52, 190}
\definecolor{ngreen}{RGB}{18, 141, 21}

\usepackage{soul}

\title{AQuilt: Weaving Logic and Self-Inspection into \\ Low-Cost, High-Relevance Data Synthesis for Specialist LLMs}

\author{Xiaopeng Ke$^{1}$~~
        Hexuan Deng$^{1}$~~
        Xuebo Liu$^{1}$\thanks{~~Corresponding Author}~~ 
        \textbf{Jun Rao}$^{1}$~~ \\
        \textbf{Zhenxi Song}$^{2}$~~ 
        \textbf{Jun Yu}$^{2}$~~ 
        \textbf{Min Zhang}$^{1}$ \\
        $^{1}$Institute of Computing and Intelligence, Harbin Institute of Technology, Shenzhen, China \\
        $^{2}$School of Intelligence Science and Engineering, Harbin Institute of Technology, Shenzhen, China \\
        \texttt{\{xiaopk7, hxuandeng, rao7jun\}@gmail.com} \\
        \texttt{\{liuxuebo,songzhenxi,yujun,zhangmin2021\}@hit.edu.cn}
  }

\begin{document}
\maketitle
\begin{abstract}
Despite the impressive performance of large language models (LLMs) in general domains, they often underperform in specialized domains. Existing approaches typically rely on data synthesis methods and yield promising results by using unlabeled data to capture domain-specific features. However, these methods either incur high computational costs or suffer from performance limitations, while also demonstrating insufficient generalization across different tasks. To address these challenges, we propose \textbf{AQuilt}, a framework for constructing instruction-tuning data for any specialized domains from corresponding unlabeled data, including \textbf{A}nswer, \textbf{Q}uestion, \textbf{U}nlabeled data, \textbf{I}nspection, \textbf{L}ogic, and \textbf{T}ask type. By incorporating logic and inspection, we encourage reasoning processes and self-inspection to enhance model performance. Moreover, customizable task instructions enable high-quality data generation for any task. As a result, we construct a dataset of 703k examples to train a powerful data synthesis model. Experiments show that AQuilt is comparable to DeepSeek-V3 while utilizing just 17\% of the production cost. Further analysis demonstrates that our generated data exhibits higher relevance to downstream tasks. Source code, models, and scripts are available at \url{https://github.com/Krueske/AQuilt}.
\end{abstract}
\section{Introduction}
\begin{figure}[t]
  \centering
  \includegraphics[width=0.97\linewidth]{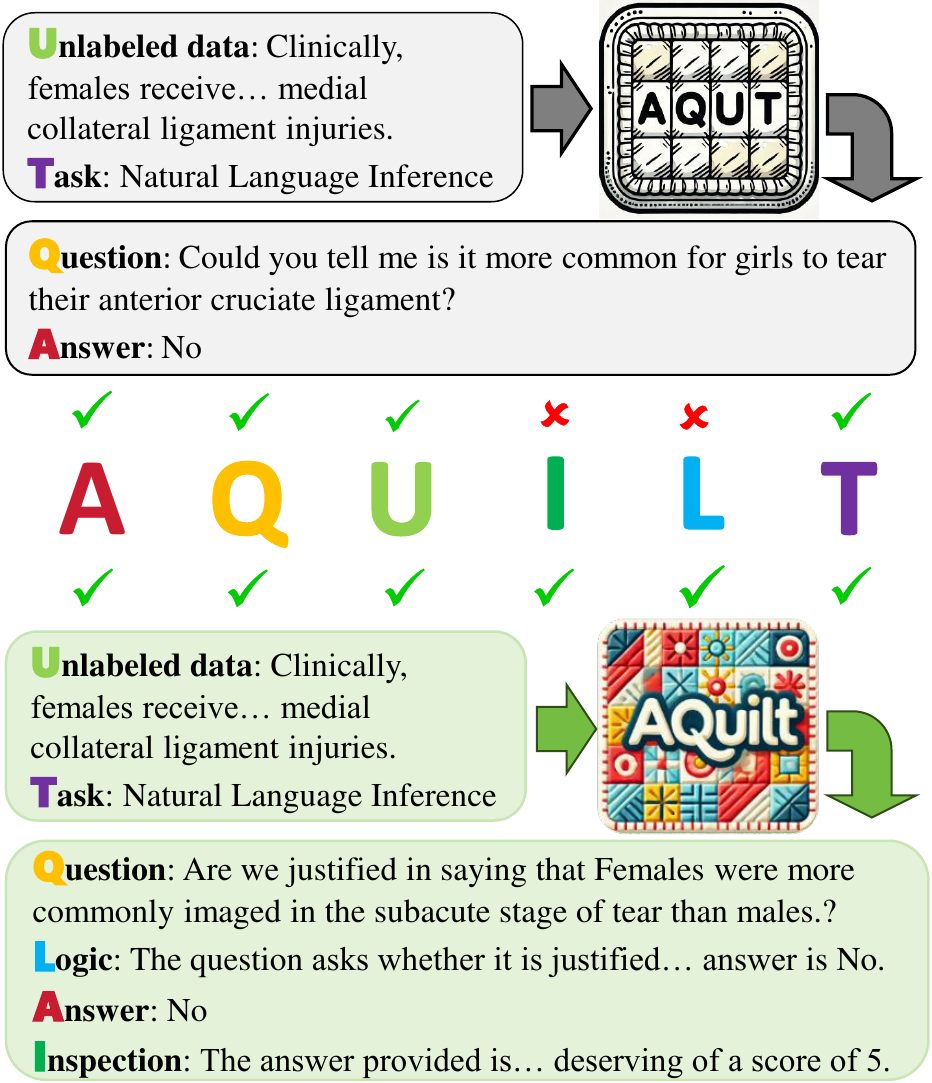}
  \caption {Traditional data synthesis models (top) can only generate question-answer pairs from unlabeled data, while AQuilt (bottom) additionally generates logic and inspection for the synthetic data.}
\end{figure}
With the rapid development of large language models (LLMs), their general capabilities have demonstrated significant success \citep{Llama3_DJP+24, OpenAIO1_JKL+24, Qwen25MathTechnical_YZH+24, DeepSeekR1Incentivizing_GYZ+25}. However, performance in specialized domains like law and medicine remains constrained \citep{DomainAdversarialTraining_GUA+16, ExposureBias_WS20}. To enhance model performance in these fields, synthetic data has emerged as a promising solution, thanks to its high-quality outputs \citep{RAFTAdapting_ZPJ+24, Phi4Technical_AAB+24, wang-etal-2024-taste, RStarMathSmall_GZL+25}.

Existing approaches rely on large models' domain priors for data synthesis. Yet, domain-specific knowledge and linguistic patterns, including lexical, syntactic, and stylistic characteristics, which are embedded in domain corpora rather than fully captured by model priors \citep{10700677}. This limitation motivates methodologies leveraging unlabeled data, which inherently encode domain-specific features \citep{hamilton-etal-2016-inducing, mudinas-etal-2018-bootstrap}. Recent advances demonstrate that unlabeled data-driven data synthesis enhances domain-specific data quality and task performance \citep{CRAFTYour_ZKES24}, supporting our focus on optimizing unlabeled data-driven methodologies.

However, although some of the methods focus on domain-specific data synthesis using unlabeled data, they still have some problems. For example, current domain synthetic data generation methods often depend on powerful commercial models or large LLMs \citep{taori2023stanford, WizardLMEmpowering_XSZ+23, AlpaGasusTraining_CLY+24}. Though they perform well, these models are usually too expensive, restricting accessibility \citep{SmallerWeaker_BHA+24}. Using smaller, specialized models is an alternative \citep{STaRBootstrapping_ZWMG22, DoGInstructPremium_CJH+24, SelfAlignmentInstruction_LYZ+24}, but their covered tasks are limited, and the generation is too simple for complex tasks.

To address limitations, we propose \textbf{AQuilt}, a framework for constructing data that incorporates \textbf{A}nswer, \textbf{Q}uestion, \textbf{U}nlabeled data, \textbf{I}nspection, \textbf{L}ogic, and \textbf{T}ask type from any unlabeled data. We train a smaller data synthesis model to synthesize domain-specific instruct-tuning data and reduce synthesis costs. We introduce \textbf{L}ogic and \textbf{I}nspection to enhance model reasoning and ensure the quality of the synthesized data \citep{STaRBootstrapping_ZWMG22, VSTaRTraining_HYM+24}. Furthermore, \textbf{T}ask type is expanded to facilitate generalization to unseen tasks during training. We then synthesize a high-quality bilingual dataset (Chinese and English) containing 703k examples using DeepSeek-V3 \citep{liu2024deepseek}, which is used to train a low-cost, high-relevance data synthesis model.

AQuilt demonstrates performance comparable to the distillation source model, DeepSeek-V3, across experiments involving two base models and five tasks, while requiring only 17\% of the production cost. Furthermore, compared to previous specialized models for data synthesis, e.g., Bonito \citep{LearningGenerate_NNTB24}, which performs well but is limited to generating data for English tasks requiring unlabeled data, our method demonstrates superior performance across these same tasks.
Further analysis confirms the effectiveness of incorporating logic and inspection, as well as the higher relevance of our synthetic data to downstream tasks, which further contributes to the model's high performance.

Our contributions are as follows:
\begin{itemize}
\item We propose AQuilt, a framework for synthesizing high-relevance data for any task from any unlabeled dataset at a low cost. By incorporating logic and inspection, we enhance model reasoning and improve data quality.
\item Experiment results show that AQuilt is comparable to DeepSeek-V3 with 17\% of the production cost.
\item Further analysis shows that logic and self-inspection contribute to better performance and more relevant generated data.
\item We will publicly release our data synthesis model, training data, and code, contributing to developing more powerful specialized LLMs and data synthesis models.
\end{itemize}

\section{Related Works}

\paragraph{Domain Data Synthesis without Unlabeled Data.}
Recent works leverage the parametric knowledge of general LLMs for domain-specific data synthesis, avoiding domain unlabeled data \citep{bao2023discmedllm, REARLReflectionAware_DJL+25,luo2025wizardmath}. Domain-oriented innovations include \citet{lawgpt} using Self-Instruct \citep{SelfInstructAligning_WKM+23} to synthesize legal question-answer pairs, and \citet{li2024scilitllm} employing GPT-4 to generate scientific questions. In particular, \citet{TinyStoriesHow_EL23} demonstrates constrained domain adaptation by generating children's stories with controlled vocabulary. While existing methods leverage strong LLMs to synthesize training data directly \citep{ChatGPTOutperforms_GAK23, MonteCarlo_XGZ+24, SelfExploreEnhancing_HKK+24}, which mainly rely on preexisting domain knowledge of commercial LLMs \citep{GPT4Technical_AAA+24, DawnLMMs_YLL+23}, their efficiency of domain data synthesis remains limited \citep{palepu2024exploring}. However, small models struggle to synthesize data with domain-specific knowledge without external input \citep{NewTermBenchmarking_DJL+24, harbola2025knowslm}, spurring research on data synthesis by combining small models with domain-specific unlabeled data.

\begin{figure*}[t]
  \centering
  \includegraphics[width=\linewidth]{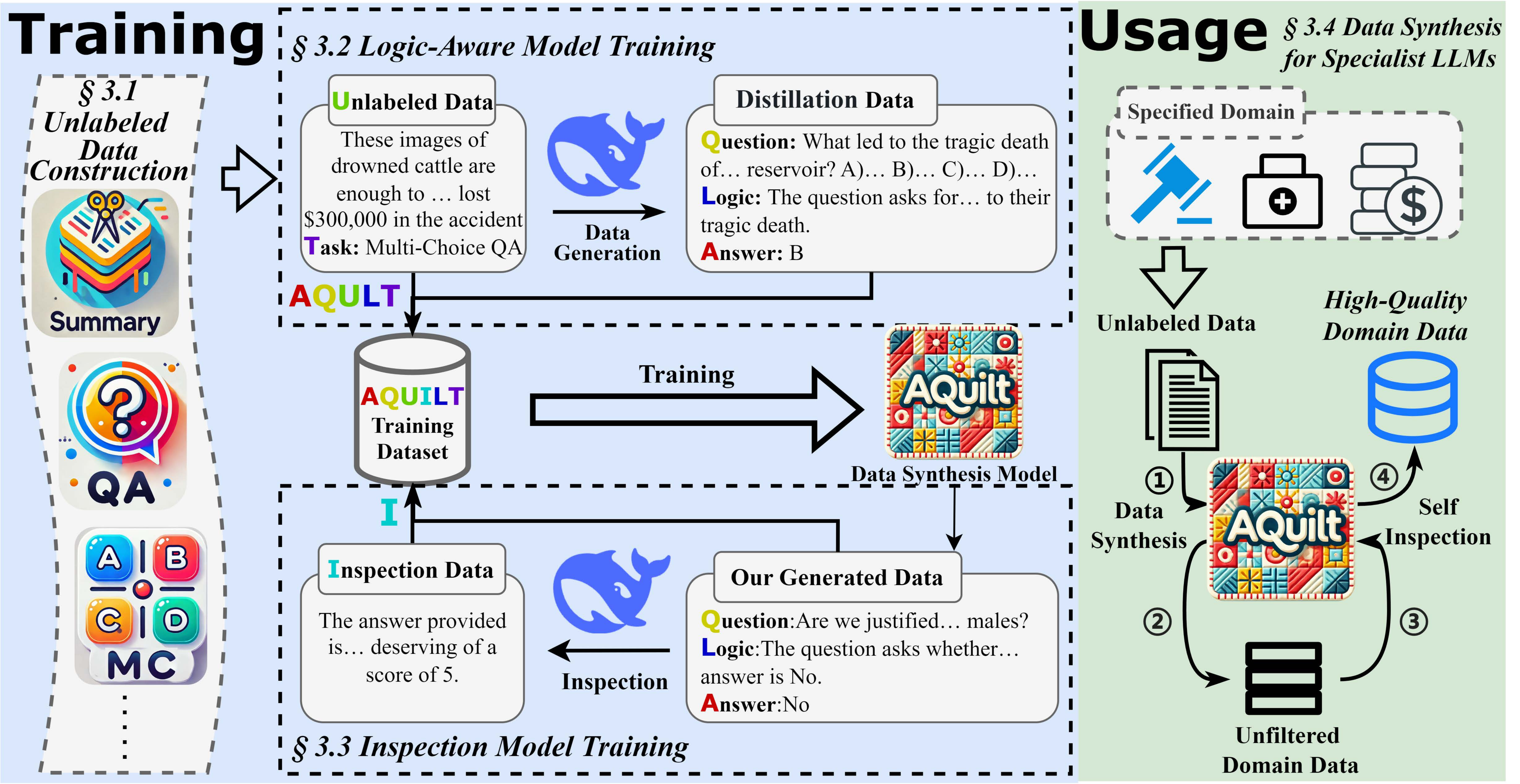}
  \caption{
  \label{fig:main}
  Overview of the proposed AQuilt framework. The left side illustrates the training process of our data synthesis model, while the right side demonstrates how the trained model automatically synthesizes high-quality domain-specific data. The synthesized data is subsequently used to train Specialist LLMs.}
\end{figure*}

\paragraph{Domain Data Synthesis with Unlabeled Data.}
Thus, we focus on domain data synthesis based on unlabeled data, which better balances performance and efficiency.
Recently, specialized methods have been proposed to integrate unlabeled data to address domain gaps \citep{bartz2022synthesis, ImprovingSimultaneous_DDL+23, 
DRPruningEfficient_DJL+25, upadhyay2025synlexlm}. For instance, \citet{LearningGenerate_NNTB24} train models on datasets with unlabeled data (e.g., summary, reading comprehension) for task-specific synthesis. \citet{CRAFTYour_ZKES24} combines retrieval with in-context learning to generate data requiring specialized knowledge. Iterative refinement techniques, such as reinforced self-training \citep{ReinforcedSelfTraining_GPS+23} and back-translation \citep{SelfAlignmentInstruction_LYZ+24}, further improve synthetic data using domain resources. However, existing solutions face two key challenges: (1) high costs and inefficiency when relying on commercial LLMs \citep{SmallerWeaker_BHA+24}, and (2) poor generalization of specialized models to out-of-distribution tasks. For example, models like \citet{DoGInstructPremium_CJH+24}, trained on GPT-generated seeds, lack the ability to define task types explicitly. These limitations underscore the urgent demand for frameworks capable of reconciling cost efficiency, domain specificity, and task generalization. In response, we present a 7B-parameter data synthesis model that integrates open-domain QA for task generalization while balancing efficiency and quality without expensive LLMs.

\section{Our Proposed AQuilt Framework}
\label{sec:aquiltframework}
To develop a low-cost, high-relevance data synthesis model with cross-task generalization capabilities for multiple domains, we propose the framework illustrated in Figure~\ref{fig:main}. It starts with building a large source data corpus (\S \ref{sec:datacollection}). 
Next, we distill \textbf{AQULT} quintuplets to enhance the model's data synthesis ability (\S \ref{sec:logic}). 
We then evaluate the generated data to obtain \textbf{I}nspection data, which further improves the model's self-inspection capability (\S \ref{sec:inspection}). 
Finally, using this synthesized data, we train a robust data synthesis model, \textbf{\text{AQuilt}}, which is applied to generate high-quality domain-specific data for training specialist LLMs (\S \ref{sec:usage}).
\begin{figure}[t]
  \centering
  \includegraphics[width=0.8\linewidth, height=0.8\linewidth]{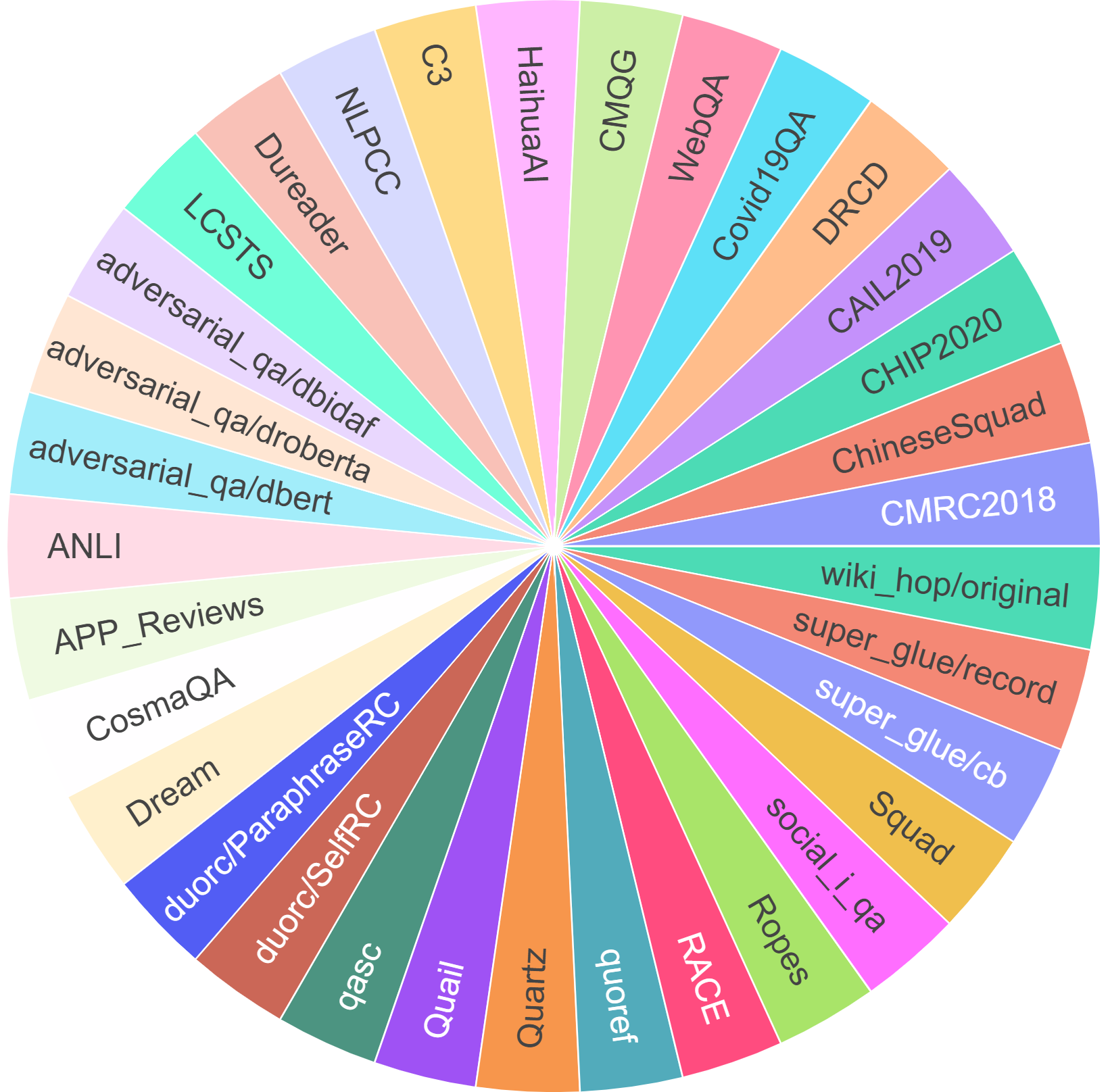}
  \caption{
  \label{fig:collect_datasets}
  Overview of the sources of the unlabeled data we collect, which covers 33 different kinds of datasets, ensuring the diversity of the data we construct.}
\end{figure}
\subsection{Unlabeled Data Construction}
\label{sec:datacollection}

\paragraph{Task Definition.} To enable our data synthesis model to synthesize various types of tasks, we follow \citet{LearningGenerate_NNTB24} and cover a wide range of tasks, including extractive QA, natural language inference, multi-choice QA (single-answer / multi-answer), text generation, text summarization, text classification, and natural language understanding. 
Furthermore, to enhance generalization across downstream tasks, we introduce two additional task types: \textbf{open-book QA} and \textbf{closed-book QA}. Because of the customizable questions, these two tasks are not confined to specific categories. When synthesizing data for a new task type, we designate it as either closed-book QA or open-book QA, depending on whether it requires unlabeled data as input. Additionally, we prepend the instruction of the new task as a prefix to the question. 
This effectively enhances the data synthesis generalization for novel domain-specific task types.

\paragraph{Data Type.} To facilitate multi-domain generalization in data synthesis, we aggregate diverse unlabeled data spanning 33 Chinese-English bilingual datasets covering news, encyclopedias, reviews, and multiple specialized domains, illustrated in Figure~\ref{fig:collect_datasets} and detailed in Appendix~\ref{sec:appendix A}. 

\subsection{Logic-Aware Model Training}
\label{sec:logic}

\paragraph{Logic-Aware Data Generation.}  
The incorporation of intermediate thought rationales has been shown to enhance LLM performance \citep{STaRBootstrapping_ZWMG22}. Motivated by this, we incorporate the model’s intermediate reasoning process, i.e., \textit{logic}, into the data synthesis procedure, fostering a more structured reasoning process and improving the overall data quality. Specifically, for each task type $t$, we randomly associate it with unlabeled data $u$ and employ a strong commercial LLM, i.e., DeepSeek-V3, to generate distilled data, including the question $q$, logic $l$, and answer $a$. Formally,

\begin{equation}
    (a, q, l) = \texttt{LLM}_{\text{Strong}}^{\text{GenData}}(u, t).
\end{equation}

Furthermore, we collect original datasets, including extractive question answering, natural language inference, multi-choice question answering (single answer), and summarization datasets from labeled data to enhance the diversity of QA pairs and address LLMs' challenges in generating extractive QA data. We further prompt the model to supplement the missing logic $l$ for these collected datasets. Formally,

\begin{equation}
    l = \texttt{LLM}_{\text{Strong}}^{\text{GenLogic}}(a, q, u, t).
\end{equation}

Collecting all synthesized data, we obtain the dataset $\mathcal{D}_L = \{(a, q, u, l, t)\}_{N}$, which comprehensively covers all the task types defined in AQuilt.

\paragraph{Relevence-Aware Data Filtering.} Existing methods, e.g., Bonito, often heavily rely on unlabeled data $u$ to generate $(q, a)$ pairs, introducing the risk of synthesizing low-relevance data for certain tasks that do not depend on unlabeled data. In contrast, we ensure that our synthesized $(q, a)$ pairs remain meaningful without $u$ in multi-choice or closed-book QA tasks that usually do not require $u$ as input. To achieve this, we explicitly guide the model’s preferences through prompt engineering. We also filter out cases not meeting this criterion by identifying prohibited words, such as ``the context'' and ``the text''. This ensures that the generated questions are applicable to downstream tasks with or without unlabeled data.

Additionally, to mitigate potential biases in LLMs \citep{ConceptualUnbiased_ZZC+24, BiasLarge_GGS+24}, we analyze word frequency statistics. For each task, we identify the most frequent words, excluding stopwords. If any word appears in more than 10\% of the data, it may indicate stylistic bias. In such cases, we eliminate questions containing these keywords, reducing their prevalence and ensuring a diverse and unbiased final training set.

After filtering, we obtain the refined dataset $\mathcal{D}^{'}_L = \{(a, q, u, l, t)\}_{N'}$. The total dataset size is summarized in Table~\ref{tab:sumdata}.

\paragraph{Model Training.} Using the large-scale dataset $\mathcal{D}^{'}_L$ obtained above, we train the data synthesis model. To enable the model to synthesize task-specific data from unlabeled data, we use $u$ and $t$ as inputs and train the model to generate $q$, $a$, and $l$. Formally:
\begin{equation}
\begin{aligned}
    \noalign{\vskip-10pt} \mathcal{L}_{\text{AQuilt}} & = -\sum_{\mathcal{D}^{'}_L}^j \log P_\theta(a_j, q_j, l_j \mid u_j, t_j). \\[-5pt]
\end{aligned}
\end{equation}

Using the above loss function $\mathcal{L}$, we obtain the data synthesis model, \(\texttt{LLM}_{\text{\text{AQuilt}}}\).

\subsection{Inspection Model Training}
\label{sec:inspection}

The above model is capable of generating data for any task from domain-specific texts. However, the generated data may still be low-quality in some situations. To mitigate this, we train the model to acquire self-inspection capabilities.

\paragraph{Inspection Data Generation.}  
To train the self-inspection capability, we need to collect training data with varying quality levels. However, since the data synthesized by strong commercial LLMs is generally of high quality, we utilize our previously trained LLM, $\texttt{LLM}_{\text{\text{AQuilt}}}$, to generate new data. This ensures that the synthesized data aligns with the distribution of our final generation process, which is beneficial for model training. Specifically, for each task $t$, we randomly sample $u$ and input it into our trained model, $\texttt{LLM}_{\text{\text{AQuilt}}}$, to synthesize data. Subsequently, we use DeepSeek-V3 to score these samples. Formally, 

\begin{equation}
\begin{aligned}
    \noalign{\vskip-15pt} (a', q', l') = & \texttt{LLM}_{\text{\text{AQuilt}}}(u, t), \\
    i = & \texttt{LLM}_{\text{Strong}}^{\text{GenInsp}}(a', q', u, l', t). \\[-10pt]
\end{aligned}
\end{equation}

As a result, we obtain the dataset $\mathcal{D}^{'}_I = \{(a', q', u, i, l', t)\}_{M}$ for training the self-inspection capability of AQuilt.

\paragraph{Self-Inspection Model Training.}
For training, we continue fine-tuning the previously trained model, $\texttt{LLM}_{\text{AQuilt}}$, by incorporating a LoRA adapter \citep{LoRALowRank_HSW+22}, formally denoted as $\texttt{LLM}_{\text{AQuilt}}^{\text{LoRA}}$. This modification enables the model to score its own instruction-tuning dataset $(a', q', l')$, which is generated based on $(u, t)$. Formally, we optimize the LoRA-augmented model using the following loss function:

\begin{equation}
\begin{aligned}
    \noalign{\vskip-15pt} \mathcal{L}_{\text{AQuilt}}^{\text{LoRA}} = - \sum_{\mathcal{D}^{'}_I}^{j} & \log P_{\theta_{\text{LoRA}}}(i_j \mid a'_j, q'_j, u_j, l'_j, t_j). \\[-10pt]
\end{aligned}
\end{equation}

\begin{table}[t]
    \centering
\scalebox{0.78}{
    \begin{tabular}{lrr}
    \toprule
    \bf Task Type & \bf English &\bf Chinese \\
    \midrule
    Extractive QA & 16k & 16k  \\
    Natural Language Inference  & 49k & 33k \\
    Multi-Choice QA (Single Answer) & 49k & 49k \\
    Multi-Choice QA (Multiple Answers)  & 28k & 31k\\
    Text Generation  & 33k & 33k\\
    Text Summarization  & 49k & 43k \\
    Text Classification & 33k & 33k \\
    Natural Language Understanding  & 32k & 32k \\
    Open-Book QA & 33k & 31k \\
    Closed-Book QA & 33k & 33k\\ \midrule
    Self-Inspection & 7k & 7k\\ \midrule
    Total & 362k & 341k \\
    \bottomrule
    \end{tabular}
}
\caption{The number of generated training data from different tasks. A total of 703k data is collected, covering both English and Chinese.}
\label{tab:sumdata}
\end{table}

\subsection{Data Synthesis for Specialist LLMs}
\label{sec:usage}
We use the constructed AQuilt model to generate high-quality domain-specific instruct-tuning data from unlabeled data. Then, we could train the specialist model on the generated data to enhance its domain-specific task performance.
\paragraph{Domain Data Synthesis.}  
When conducting downstream task learning for LLMs, we synthesize high-quality domain-specific data using only the specified task type $t$ and the relevant domain unlabeled data $u$. Leveraging our data synthesis model, we can efficiently generate training data tailored to the given domain and task. Formally, 

\begin{equation}
\begin{aligned}
    \noalign{\vskip-15pt} (a', q', l') = & \texttt{LLM}_{\text{AQuilt}}(u, t). \\[-10pt]
\end{aligned}
\end{equation}

Notably, if the task type $t$ has not been observed in the training set, we designate $t$ as either \textit{closed-book QA} or \textit{open-book QA}, depending on whether it requires the unlabeled data as input. Additionally, we prepend the instruction of the new task as a prefix to the question.

\paragraph{Data Self-Inspection.}  
To ensure the quality of the generated data, we apply filtering based on self-inspection. Specifically, we first generate an inspection score using the trained model:

\begin{equation}
\begin{aligned}
    i' = & \texttt{LLM}_{\text{AQuilt}}^{\text{LoRA}}(a', q', u, l', t).
\end{aligned}
\end{equation}

Subsequently, we filter out low-quality data. By default, we remove data with an inspection score of 2 or lower (on a 5-point scale). If more than 20\% of the data receives a score of 2, which suggests that the task is inherently simpler, we only remove data with a score of 1. As a result, we obtain high-quality training data, facilitating efficient adaptation of the model to specific domains.
\paragraph{Training Specialist LLM}
We train the target model on high-quality domain-specific data synthesized by AQuilt model to enhance its performance on domain-specific tasks.
\section{Experiment}
\label{section5}

\begin{table*}[h!]
\centering
\scalebox{0.695}{
\begin{tabular}{llrrrrrrrrrrrr}
\toprule
 \multirow{2}{*}{\bf Model} & \multirow{2}{*}{\bf Source} & \multicolumn{2}{c}{\textbf{SquadQA}} & \multicolumn{2}{c}{\textbf{PubMedQA}} & \multicolumn{2}{c}{\textbf{CEVAL}} & \multicolumn{2}{c}{\textbf{Translation}} & \multicolumn{2}{c}{\textbf{EssayQA}} & \multicolumn{2}{c}{\textbf{Avg.}} \\
 \cmidrule(lr){3-4} \cmidrule(lr){5-6} \cmidrule(lr){7-8} \cmidrule(lr){9-10} \cmidrule(lr){11-12} \cmidrule(lr){13-14} 
 & & \bf Score & \bf Cost & \bf Score & \bf Cost  & \bf Score & \bf Cost & \bf Score & \bf Cost  & \bf Score & \bf Cost & \bf Score & \bf Cost \\
 \midrule
\multirow{7}{*}{\rotatebox{90}{\textbf{Qwen2.5-7B}}}
& \bf None & 3.12 & 0 & 56.60 & 0 & 87.46 & 0 & 32.23 & 0 & 19.21 & 0 & 39.72 & 0\\
& \bf TAPT & 3.16 & 0.90 & 56.40 & 1.13 & 87.72 & 2.32 & 33.00 & 1.88 & 19.11 & 1.61 & 39.88 & 1.57\\
& \bf Bonito & 22.78 & 1.19 & 71.40 & 1.42 & NA & NA & NA & NA & NA & NA & NA & NA\\
& \bf DeepSeeek-V3 w/ Self-Instruct & NA & NA & 74.00 & 10.40 & 87.81 & 16.73 & \underline{36.95} & 20.31 & \bf 24.07 & 26.02 & NA & NA\\
& \bf DeepSeeek-V3 w/ Unlabeled Data & 16.09 &3.91 & \underline{75.80} & 4.65 & \bf 88.55 & 7.21 & 36.89 & 9.14 & 20.73 & 12.96 & 47.61 & 7.57\\
& \bf DeepSeeek-V3 w/ SI+UD & \underline{30.20} & 6.33 & \bf 76.80 & 7.55 & 88.34 & 12.88 & 36.81 & 18.32 & \underline{23.22} & 24.47 & \bf 51.47 & 13.91 \\
& \bf AQuilt & \bf 34.69 & 1.48 &  74.00 & 1.75 &  \underline{88.44} & 2.90 & \bf 38.00 & 2.44 & 22.11 & 2.25 & \underline{51.45} & 2.16\\ \midrule
\multirow{7}{*}{\rotatebox{90}{\textbf{Llama3-8B}}}
& \bf None & 3.68 & 0 & 73.60 & 0 & 58.32 & 0 & 27.79 & 0 & 15.37 & 0 & 35.75 & 0 \\
& \bf TAPT & 3.67 & 1.22 & 73.60 & 1.56 & 58.50 & 3.73 & 28.13 & 3.07 & 15.20 & 2.38 & 35.82 & 2.39\\
& \bf Bonito & 23.05 & 1.51 & 72.20 & 1.85 & NA & NA & NA & NA & NA &NA & NA & NA\\
& \bf DeepSeeek-V3 w/ Self-Instruct & NA & NA & 74.80 & 10.75 &  61.96 & 17.95 & 34.07 & 21.50  & \underline{21.26} & 26.57 & NA & NA\\
& \bf DeepSeeek-V3 w/ Unlabeled Data & 16.69 &4.23 & \underline{75.80} & 4.91 & 59.25 & 8.43 & \underline{34.67} & 10.33 & 19.27 & 13.52 & 41.14 & 8.28\\
& \bf DeepSeeek-V3 w/ SI+UD & \underline{32.01} & 6.65 & \bf 76.40 & 7.81 & \bf 64.16 & 14.10 & \bf35.57 & 19.51 & \bf 21.93 & 25.03 & \underline{46.01} & 14.62\\
& \bf AQuilt& \bf 40.89 & 1.79 &  75.20 & 2.10 &  \underline{63.16} & 3.91 & 34.50 & 3.35 &  19.65 & 2.77 & \bf 46.68 & 2.78\\
\bottomrule
\end{tabular}
}
\caption{Main results for downstream task learning. ``Model'' indicates the base models used for training, all of which are the instruction versions. ``Source'' represents the source of the training data synthesis, detailed in \S \ref{sec:setup}. ``Cost'' represents the total expense for data synthesis and training. ``Avg.'' represents the average scores across all tasks. \textbf{Bold} indicates the best performance, while \underline{underline} indicates the second-best for each task.}
\label{tab:results_Instruct}
\end{table*}
\subsection{Training Setup of AQuilt}
\label{sec:aquiltraining}
\paragraph{Data.} As summarized in Table~\ref{tab:sumdata}, we construct a bilingual data set (EN / ZH) that covers 10 types of tasks, with the aim of improving model generalization through diversified coverage. To prevent task dominance and ensure balanced distributions, we apply downsampling: logic training data are capped at 50k samples per (task, language) pair, while self-inspection data contain no more than 2k samples per score for each language, with scores ranging from 1 to 5. The final aggregated dataset comprises 703k samples, with full prompts detailed in Appendix~\ref{sec:appendix C}.

\paragraph{Training.} The experiments are conducted on Qwen2.5-7B-Base with 8 NVIDIA 4090 24GB GPUs. For both training procedures described above, we use the AdamW optimizer, with a learning rate of 1e-4, batch size of 32, and LoRA r and alpha both set to 64, training for 2 epochs.

\subsection{Evaluation Setup of AQuilt}
\label{sec:setup}

\paragraph{Benchmark.}
In this work, we conduct experiments across distinct downstream tasks, covering various task types. For extractive QA, we use \textbf{SquadQA} \citep{KnowWhat_RJL18} and follow the online adaptation setting \citep{hu2023metalearning}, which teaches LLM domain knowledge contained in unlabeled data. For yes/no QA, we select \textbf{PubMedQA} \citep{jin2019pubmedqa}, an English natural language inference task related to medical research papers. For multi-choice QA, we choose eight subjects from \textbf{CEVAL} \citep{huang2023ceval}, covering compulsory courses from middle school to university in China. For translation and open-ended QA, we utilize the Legal \textbf{Translation} and Legal \textbf{EssayQA} tasks from LexEval \citep{li2024lexevalcomprehensivechineselegal}. These tasks span different domains, validating the cross-domain and cross-task capabilities of our data synthesis model.

We evaluate PubMedQA and CEVAL using accuracy. Following \citet{KnowWhat_RJL18}, we use the SQuAD F1 score to evaluate the SquadQA test dataset. For the Translation and EssayQA tasks, as in LexEval \citep{li2024lexevalcomprehensivechineselegal}, we compute the Rouge-L score on the generated output. Also, we compute the BERTScore for the Translation and EssayQA tasks, listed in the Appendix~\ref{sec:appendix B}, to ensure the robustness of the evaluation metrics.

\paragraph{Domain Data Generation.}
AQuilt requires domain-specific unlabeled data, which we source as follows: SquadQA uses test set data \citep{hu2023metalearning}, PubMedQA uses its original training set, CEVAL collects textbooks, while legal tasks (Translation and EssayQA) use Chinese CAIL \citep{cail2024} and English MAUD/UK-Absp \citep{wang2023maud,bhattacharya2021} datasets. Task-specific data is generated following \S \ref{sec:usage}, with unseen tasks like Translation and EssayQA framed as closed-book QA via question prefixes. With vLLM \citep{kwon2023efficient} (temperature=0.7, top\_p=0.95, max\_length=1024), we synthesize 20k training samples per task.

\paragraph{Baselines.}
We compare different baselines based on the source of synthetic training data. For the ``None'' baseline, we directly prompt the model for evaluation without any training. For the ``TAPT'' baseline, we follow \citet{gururangan-etal-2020-dont} to use task-adaptive pretraining, training the model only on unlabeled data. For the ``Bonito'' and ``DeepSeek-V3 (w/ unlabeled data)'' baselines, we use these models to synthesize data based on domain-specific texts and tasks, then fine-tune the models on the synthesized data. For the ``DeepSeek-V3 (w/ Self-Instruct)'' baseline, we generate data by applying Self-Instruct \citep{SelfInstructAligning_WKM+23} (a method enabling model to autonomously create training instructions) to DeepSeek-V3 and adapt the prompt from SELF-GUIDE \citep{zhao2024selfguide}  for the synthesis of domain-specific task data. For the ``DeepSeek-V3 (w/ Self-Instruct + Unlabeled Data)'' baseline, which is abbreviated as ``DeepSeek-V3 (w/ SI + UD)'' in Table~\ref{tab:results_Instruct}, we incorporate unlabeled data into the Self-Instruct data synthesis process (Specific prompts in Appendix~\ref{sec:appendix C}).

\paragraph{Training Details for Specialist LLMs.}
To validate the effect of the synthetic data, we perform experiments based on instruct models, including Qwen2.5-7B-Instruct \citep{qwen2.5} and Llama3-8B-Instruct \citep{Llama3_DJP+24}. We fine-tune these models with LoRA on different supervision sources for 3 epochs per task. All other settings are consistent with those in Section~\ref{sec:sec:aquiltraining}. Note that we use a lower LR of 1e-7 and single-epoch training for CEVAL based on Qwen2.5 (prone to overfitting given Qwen2.5's strong baseline) and TAPT (to preserve instruction-following capability). For TAPT specifically, we reformat tasks into tuning prompts like ``Please output [Domain] text in English: [Sentence]'' to maintain alignment.

\subsection{Main Results}

\paragraph{Superior Performance.} As shown in Table \ref{tab:results_Instruct}, AQuilt outperforms most baselines on average and is comparable to the best setting using DeepSeek-V3 (w/ SI + UD). Besides, TAPT, which relies solely on unlabeled data, shows no significant improvement across tasks, highlighting the value of labeled data synthesis. SquadQA tests whether synthetic data enables effective learning of domain-specific knowledge from unlabeled data. Since DeepSeek-V3 (w/ Self-Instruct) lacks unlabeled data, it cannot follow this setup, hence its results are marked as NA. In contrast, AQuilt significantly improves results on this task, confirming LLMs' poor performance on extractive QA tasks and justifying the use of original QA from labeled data (introduced in \S \ref{sec:logic}) for such tasks.

\paragraph{Low-Cost Generation.} We compute the cost of different data synthesis and training methods in dollars. Given the DeepSeek-V3 model's 671B size makes local deployment impractical, we use its official API and base the cost calculation on total data synthesis expenditure. For other sources, we instead use local NVIDIA 4090 24GB GPUs, calculating costs via Vast AI\footnote{\url{https://vast.ai/}}'s GPU rental prices. For production costs, we calculate the total cost based on the GPU Hours used.

AQuilt matches DeepSeek-V3 (w/ SI + UD) in performance at 17\% of its cost, and achieves better results than DeepSeek-V3 (w/ Unlabeled Data) using 31\% of the cost, demonstrating AQuilt's significant efficiency advantages in data synthesis.

\paragraph{Cross-Task Generalization.} Bonito is unable to generate data for three tasks (results are marked as NA) due to its support for only English tasks that rely on unlabeled data. In contrast, AQuilt achieves better task generalization by incorporating Chinese data and defining more flexible task types. 
Specifically, AQuilt assigns the task type as closed/open-book QA and uses task requirements as question prefixes. 
Experimental results demonstrate that AQuilt consistently outperforms various baselines on these tasks, highlighting its strong ability to generalize across different tasks.

\paragraph{Effectiveness of Unlabeled Data.} By comparing DeepSeek-V3 (w/ SI + UD) with DeepSeek-V3 (w/ Self-Instruct), it's evident that even on the strong DeepSeek-V3 model foundation, incorporating unlabeled data during domain data synthesis can further improve synthetic data quality. This highlights the effectiveness of using domain-specific unlabeled data for data synthesis.

\section{Analysis}

We provide a comprehensive analysis to demonstrate the effect of our method and the underlying reasons for its success. Given the implementation cost, unless otherwise specified, the experiments are based solely on the results obtained from Llama3-8B-Instruct as the base model and focus on SquadQA, CEVAL, and Translation.

\begin{table}[t]
\centering
\scalebox{0.68}{
\begin{tabular}{lccccc}
\toprule
 \bf Model & \bf SquadQA & \bf CEVAL & \bf Translation & \bf Avg.\\
 \midrule
 \bf AQuilt & \bf 40.89 & \bf 63.16 & \bf 34.50 & \bf 46.18 \\
  \quad \bf w/o Logic & 40.68 & 59.64 & 33.61 & 44.64 \\ 
  \quad \bf w/o Self-Inspection & 40.00 & 60.95 & 34.22 & 45.06 \\ 
  \quad \bf w/ Low-Quality& 39.81 & 59.22 & 33.15 & 44.06 \\
\bottomrule
\end{tabular}
}
\caption{Ablation results for logic and self-inspection. We independently remove logic and self-inspection to observe performance changes. w/ Low-Quality refers to using self-inspection to select low-quality data, in contrast to the main experiment settings.}
\label{tab:results_ablation}
\end{table}

\subsection{Analysis on Logic and Self-Inspection}

To evaluate the effects of logic and self-inspection, we independently remove each component during the training process. Results are in Table \ref{tab:results_ablation}.

\paragraph{Effect of Logic.} To validate the effect of logic, we remove the logic component from the entire training pipeline (w/o Logic) and retrain AQuilt using data without logic, while maintaining all other settings and experimental setups. In subsequent data synthesis, no logic is incorporated. The results show a significant drop in model performance after removing logic, demonstrating its critical role within the entire data synthesis framework.

\paragraph{Effect of Self-Inspection.} To assess the impact of self-inspection, we conduct the following evaluation: the optimal setting for filtering low-quality data (AQuilt), a no-filtering setup (w/o Self-Inspection), and the use of only low-quality data identified by self-inspection (w/ Low-Quality). To ensure comparability, consistent data volumes are applied across all settings. The results confirm the significant effect of self-inspection. Further, under w/ Low-Quality, we observe a slight decrease in performance, indicating that the presence of low-quality data negatively impacts model performance. Fortunately, when logic is applied, overall performance remains at a relatively high level.

\subsection{Domain Relevance of Generated Data}

We demonstrate that our method achieves superior performance by generating data with higher relevance to the target domain and lower noise. To demonstrate this, we select CEVAL, Translation, and SquadQA as three distinct domain tasks, each with 2k synthetic data samples. Using the Qwen2.5-7B, we compute sentence embeddings for the synthetic questions generated by DeepSeek-V3 (w/ Unlabeled Data) and AQuilt, which are both created based on unlabeled data.

\begin{figure}[t]
   \centering
   \begin{minipage}[t]{.49\linewidth} 
      \centering
      \includegraphics[width=\linewidth]{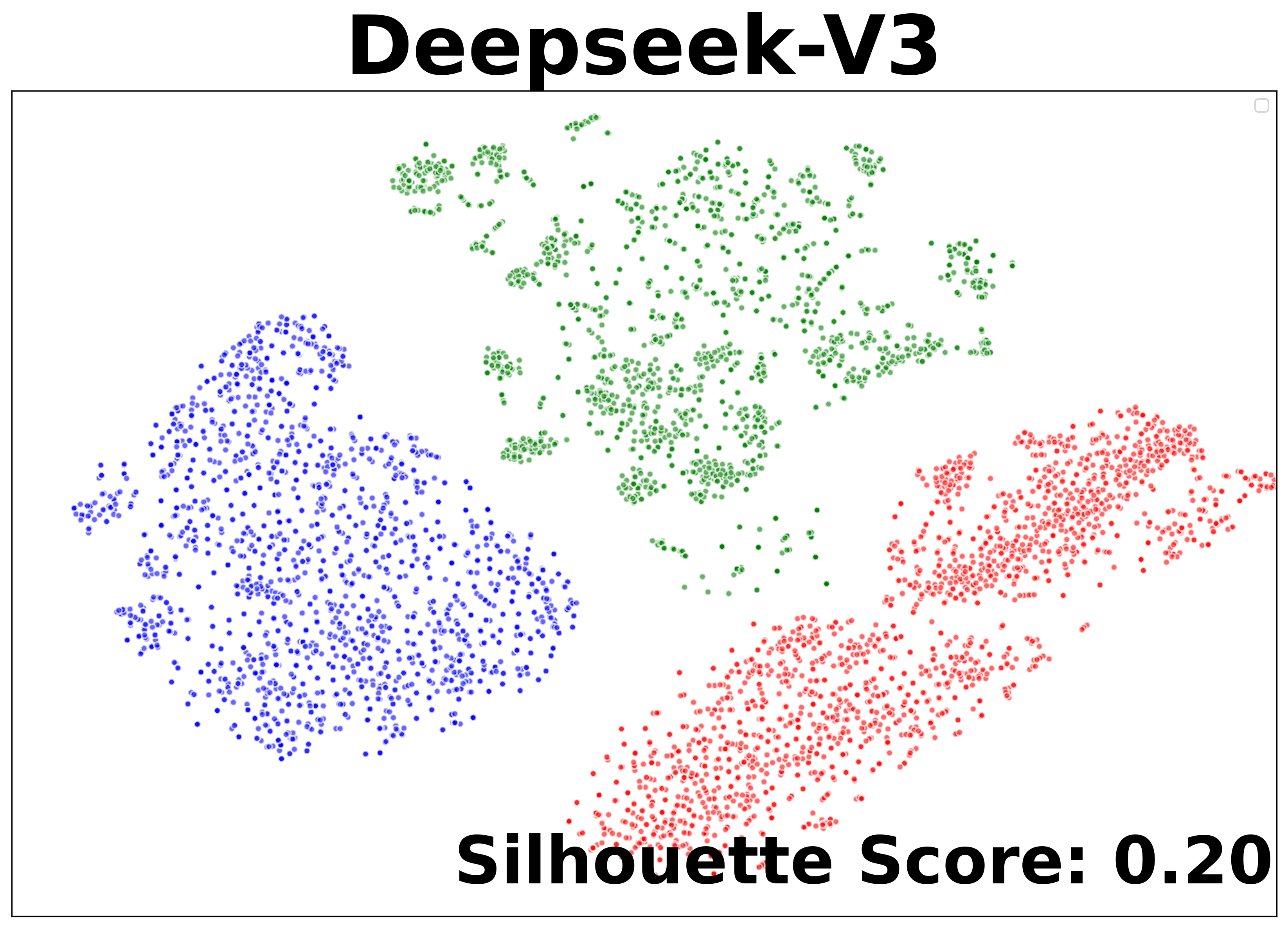}
   \end{minipage}
   \begin{minipage}[t]{.49\linewidth} 
      \centering
      \includegraphics[width=\linewidth]{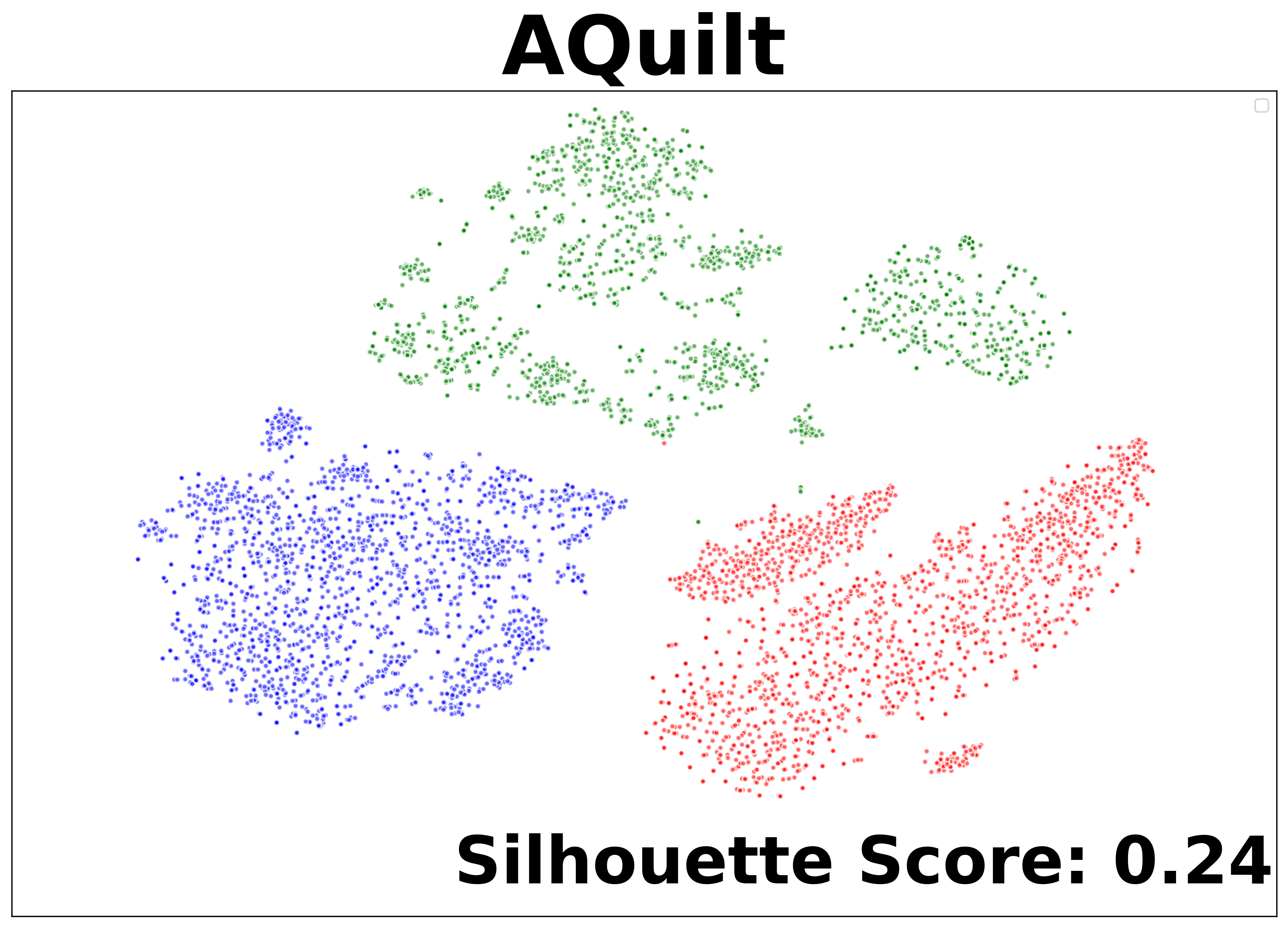}
   \end{minipage}
   \caption{
   \label{fig:tsne}
   Relevance analysis of synthesized domain data. We convert generated questions into sentence vectors and analyze the distribution across CEVAL (Red), Translation (Green), and SquadQA (Purple).}
\end{figure}

As shown in Figure~\ref{fig:tsne}, we apply t-SNE for dimensionality reduction and plot a 2D scatter plot. The results show that the generated data of AQuilt is more concentrated and contains fewer noises. To confirm this quantitatively, we compute the Silhouette Score, which reflects consistency and relevance within data clusters, with higher values indicating stronger domain relevance. The results align with the scatter plot, demonstrating that our method generates more concentrated data with reduced noise and increased relevance.

\begin{table}[t]
    \centering
\scalebox{0.7}{
    \begin{tabular}{lcccc}
    \toprule
    \bf Model & \bf SquadQA & \bf CEVAL & \bf Translation & \bf Avg. \\
    \midrule
   \bf DeepSeek-V3  & 6.90\% &  8.18\% &  0.00\% &  5.23\%\\
   \bf AQuilt  & \bf 0.40\% & \bf 5.15\% &  0.00\% &  \bf 1.85\%\\
   \bottomrule
    \end{tabular}
    }
    \caption{Independence analysis of synthetic data with unlabeled data. We assess the percentage of synthetic questions generated by DeepSeek-V3 and AQuilt that are highly dependent on unlabeled data, revealing their low relevance to downstream tasks.}
    \label{tab:independence_analysis}
\end{table}

\begin{table}[t]
\centering
\scalebox{0.65}{
\begin{tabular}{lccccr}
\toprule
 \multirow{2}{*}{\bf Source}
 & \multirow{2}{*}{\bf SquadQA} & \multirow{2}{*}{\bf CEVAL} & \multirow{2}{*}{\bf Translation } & \multicolumn{2}{c}{\bf Avg. } \\
  \cmidrule(lr){5-6} 
 &   &  &   & \bf Score & \bf Cost \\
 \midrule
 \bf Qwen2.5-72B  & 21.19  & 59.06  & \bf 34.82  & 38.36 & 19.92$\times$\\ 
 \bf AQuilt& \bf 40.89  & \bf 63.16  & 34.50 & \bf 46.18 & \bf 1$\mathbf{\times}$\\
\bottomrule
\end{tabular}
}
\caption{Comparison with Qwen2.5-72B-Instruct. All abbreviations are consistent with those in Table~\ref{tab:results_Instruct}.}
\label{tab:compare_with_qwen72B}
\end{table}

\begin{table}[t]
    \centering
\scalebox{0.8}{
    \begin{tabular}{lcccc}
    \toprule
    \bf Model & \bf SquadQA & \bf CEVAL & \bf Translation & \bf Avg. \\
    \midrule
   \bf Bonito  & 2.43 & NA  & NA  & NA \\
   \bf AQuilt  & 2.98 & 4.16 &  3.58 & 3.57\\
   \bottomrule
    \end{tabular}
    }
    \caption{Using GPT-4o to evaluate 1,000 samples randomly drawn from the training data synthesized by AQuilt and Bonito for the three test tasks.}
    \label{tab:GPT-4o_evaluation}
\end{table}

\subsection{Analysis of Relevance-Aware Filtering}

For multi-choice and closed-book QA tasks, the synthetic labeled data must remain independent of the unlabeled data; otherwise, introducing false correlations during training may lead to low relevance to these tasks. To address this, we apply relevance-aware data filtering (introduced in Section~\ref{sec:logic}).

In the unlabeled data-based synthetic generation setting (Table \ref{tab:results_Instruct}), while AQuilt demonstrates consistent advantages across tasks, DeepSeek-V3 (w/ Unlabeled Data) shows curiously limited gains on CEVAL. To analyze the underlying cause, we examine the proportion of generated questions that rely on unlabeled data to generate answers, which may contribute to hallucinations. We sample 2,000 instances from the generated dataset and use GPT-4o to evaluate. As shown in Table \ref{tab:independence_analysis}, the results indicate that even when explicitly instructed in the prompt, strong models like DeepSeek-V3 still tend to generate questions with spurious correlations, resulting in low relevance to downstream tasks and consequently reducing overall performance.

\subsection{Ablation of Base Model}

To ensure a fair comparison of improvement sources, we conduct controlled experiments under identical settings (based on the same unlabeled data) between AQuilt (trained on Qwen2.5-7B) and the Qwen family's larger 72B model, confirming our enhancements originate from methodology rather than base model capacity.

As shown in Table \ref{tab:compare_with_qwen72B}, our method outperforms the 72B model on average, while requiring only about 1/20 GPU hours on NVIDIA A800 80G. The overall experimental results align with the main experimental trends, further validating the effect of our method and confirming that the superior performance is not due to a stronger base model.

\subsection{GPT-4o Evaluation}
To validate the quality of the questions generated by AQuilt and to compare its performance with models of comparable scale, Bonito, we present the scoring results (without Self-Inspection filtering) obtained in Table \ref{tab:GPT-4o_evaluation} by using GPT-4o (temperature=0.7, top\_p=0.95) to evaluate 1,000 samples randomly drawn from the training data synthesized by AQuilt and Bonito for each task. The prompts used by GPT-4o are completely consistent with those shown in the Appendix~\ref{sec:appendix C}, with a scoring scale ranging from 1 to 5 points.

Based on the prompt we provide to GPT-4o, a score of 2 points meets the basic quality requirement. As seen in Table \ref{tab:GPT-4o_evaluation}, the average score for AQuilt-synthesized data across most tasks exceed 3 points. However, Bonito-synthesized data receive lower score. Meanwhile, since Bonito could not synthesize data for the CEVAL and Translation tasks, these are marked as NA. This indicates that the majority of data synthesized by AQuilt is relatively high-quality. For comparatively simpler tasks such as SquadQA and Translation, GPT-4o tends to assign slightly lower scores, demonstrating that the difficulty of domain-specific tasks can influence the final evaluation results to some extent.

\section{Conclusion}

In this paper, we present AQuilt, a framework for generating data that incorporates \textbf{I}nspection, \textbf{Q}uestion, \textbf{U}nlabeled data, \textbf{A}nswer, \textbf{L}ogic, and \textbf{T}ask type from unlabeled data. Specifically, AQuilt enhances data synthesis quality through the introduction of \textbf{L}ogic and \textbf{I}nspection. The inclusion of \textbf{T}ask type, encompassing both open-book QA and closed-book QA, enables cross-task generalization for downstream data synthesis. Experimental results demonstrate that AQuilt outperforms Bonito, a widely used data synthesis model, in both task generalization and performance. Our synthetic data is even comparable to that of DeepSeek-V3, while requiring less than 17\% of the production cost. Further analysis reveals that while the use of open-source LLMs yields generally favorable results, it shows drawbacks in format adherence and downstream task relevance. This underscores the necessity of dedicated data synthesis models. We will release all the training details and models to encourage further research.

\section*{Limitations}

\paragraph{Additional Data Synthesis Sources.} In this work, we used only DeepSeek-V3 as a source of distilled data. Expanding the data sources to include human-curated existing training datasets and data synthesized by more powerful models could provide a diverse mix of training data. This expansion may further enhance the diversity of the synthesized data styles, as well as improve downstream model performance and robustness.

\paragraph{Data Synthesis for Various Languages.} In this work, we extend the language capabilities of the data synthesis model to two high-resource languages: Chinese and English. In future work, we are interested in exploring the model's performance on mid- to low-resource languages, where the model may have poorer performance and less data availability. Additionally, we will investigate whether the model exhibits zero-shot generalization capabilities when encountering languages not seen during training.

\paragraph{Advanced Data Synthesis Frameworks.} With the introduction of DeepSeek-R1 \citep{DeepSeekR1Incentivizing_GYZ+25} and Kimi-K1.5 \citep{KimiK15_DGX+25}, more advanced data synthesis frameworks have emerged, which use iterative data synthesis, high-quality evaluation, and reinforcement learning to generate progressively stronger data. Our self-inspection training framework has the potential to generalize to these frameworks. We are interested in exploring this setting in future work.

\section*{Ethics Statement}
Our work adheres to the ACL Ethics Policy and publicly released the code for reproducibility. LLMs may exhibit racial and gender biases, so we strongly recommend users assess potential biases before applying the models in specific contexts. Additionally, due to the difficulty of controlling LLM outputs, users should be cautious of issues arising from hallucinations.

\section*{Acknowledgments}
This work was supported in part by the National Natural Science Foundation of China (Grant No. 62206076), Guangdong S\&T Program (Grant No. 2024B0101050003), Guangdong Basic and Applied Basic Research Foundation (Grant No. 2024A1515011491), and Shenzhen Science and Technology Program (Grant Nos. ZDSYS20230626091203008, KJZD20231023094700001, KQTD20240729102154066). 
We would like to thank the anonymous reviewers and meta-reviewer for their insightful suggestions.

\bibliography{custom,add}
\clearpage
\appendix

\section{Unlabeled Data Construction Details}
\label{sec:appendix A}

We introduce the sources of the collected datasets, including unlabeled data and labeled data dataset, in which we collect $(u, q, a, t)$ tuples to ensure higher quality.

\paragraph{Sources of Unlabeled Data.} We collect unlabeled data from the following 33 datasets: CMRC2018 \citep{cui-etal-2019-span}, ChineseSquad \citep{ChineseSquad}, CHIP2020 \citep{tianchi_competition_2020}, CAIL2019 \citep{cail2019}, DRCD \citep{chen2018drcd}, Covid19QA \citep{datafountain_covid19_qa}, WebQA \citep{kexuefm-4338}, CMQG \citep{tianchi2024tcmqg}, HaihuaAI \citep{haihua_ai_competition_2021}, C3 \citep{sun2019investigating}, NLPCC \citep{nlpcc2017}, Dureader \citep{dureader2018}, LCSTS \citep{chen2015lcsts}, AdversarialDbidaf, AdversarialDroberta, AdversarialDbert \citep{bartolo2020beat}, ANLI \citep{nie-etal-2020-adversarial}, APPReviews \citep{sealuzh_user_quality}, CosmaQA \citep{huang2019cosmosqa}, Dream \citep{sundream2018},  Duorc \citep{saha-etal-2018-duorc}, Qasc \citep{khot2020qasc}, Quail \citep{rogers2020getting}, Quartz \citep{tafjord2019quartz}, Quoref \citep{dasigi-etal-2019-quoref}, RACE \citep{lai2018race}, Ropes \citep{lin-etal-2019-reasoning}, SocialQA \citep{mihaylov2019social}, Squad \citep{rajpurkar2016squad},  SuperGLUE \citep{wang2019superglue},  Record \citep{choi2019record}, WikiHop \citep{berant2018wikihop}.

\paragraph{Sources of Labeled Data.} We collect labeled data from the following 21 datasets: CMRC2018 \citep{cui-etal-2019-span}, ChineseSquad \citep{ChineseSquad}, CHIP2020 \citep{tianchi_competition_2020}, CAIL2019 \citep{cail2019}, HaihuaAI \citep{haihua_ai_competition_2021}, C3 \citep{sun2019investigating}, DRCD \citep{chen2018drcd}, Covid19QA \citep{datafountain_covid19_qa}, WebQA \citep{kexuefm-4338}, CMQG \citep{tianchi2024tcmqg}, Dureader \citep{dureader2018}, LCSTS \citep{chen2015lcsts}, AdversarialDbidaf, AdversarialDroberta, AdversarialDbert \citep{bartolo2020beat}, ParaphraseRC, SelfRC \citep{saha-etal-2018-duorc}, Quoref \citep{dasigi-etal-2019-quoref}, Ropes \citep{lin-etal-2019-reasoning}, Squad \citep{rajpurkar2016squad}, Record \citep{choi2019record}.

\section{Translation and EssayQA Task Evaluation Results}
\label{sec:appendix B}
In the Table~\ref{tab:results_BertScore}, we show the results of Translation and EssayQA tasks evaluated by Rouge-L and BERTScore. It can be seen that the data synthesized by AQuilt significantly improves the performance of the Instruct model on these tasks, with improvements comparable to DeepSeek-V3.
\begin{table*}[t]
\centering
\scalebox{0.8}{
\begin{tabular}{llrrrr}
\toprule
 \multirow{2}{*}{\bf Model} & \multirow{2}{*}{\bf Source} & \multicolumn{2}{c}{\textbf{Translation}} & \multicolumn{2}{c}{\textbf{EssayQA}} \\
 \cmidrule(lr){3-4} \cmidrule(lr){5-6} 
 & & \bf Rouge-L & \bf BERTScore & \bf Rouge-L & \bf BERTScore \\
 \midrule
\multirow{6}{*}{\rotatebox{0}{\bf Qwen2.5-7B}}
& \bf None &  32.23 & 91.53 & 19.21 & 86.27 \\
& \bf TAPT & 33.00 & 91.52 & 19.11 & 86.27 \\
& \bf Bonito & NA & NA & NA & NA \\
& \bf DeepSeek-V3 w/ Self-Instruct & \underline{36.95} & 93.10 & \bf 24.07 & \bf 87.30 \\
& \bf DeepSeek-V3 w/ Unlabeled Data &36.89 & 92.95 & 20.73 & 86.52 \\
& \bf DeepSeek-V3 w/ SI + UD & 36.81 & \underline{93.15} & \underline{21.93} & \underline{86.96} \\
& \bf AQuilt &\bf 38.00 & \bf 93.46 & 22.11 & 86.52\\ \midrule
\multirow{6}{*}{\rotatebox{0}{\bf Llama3-8B}}
& \bf None & 27.79 & 89.94 & 15.37 & 82.51  \\
& \bf TAPT & 28.13 & 91.62 & 15.20 & 82.46\\
& \bf Bonito & NA & NA & NA &NA \\
& \bf DeepSeek-V3 w/ Self-Instruct & 34.07 & 92.94  & \underline{21.26} & \underline{86.96}\\
& \bf DeepSeek-V3 w/ Unlabeled Data &  \underline{34.67} &  \underline{93.06} & 19.27 & 86.29 \\
& \bf DeepSeek-V3 w/ SI + UD & \bf 35.57 & \bf 93.13 & \bf 23.22 & \bf 87.02 \\
& \bf AQuilt  & 34.50 & 92.97 & 19.65 & 86.18 \\
\bottomrule
\end{tabular}
}
\caption{The evaluation results of Translation and EssayQA tasks using Rouge-L and BERTScore as evaluation metrics respectively. \textbf{Bold} indicates the best performance, while \underline{underline} indicates the second-best for each task.}
\label{tab:results_BertScore}
\end{table*}

\section{Prompts}
\label{sec:appendix C}

We present all the prompts used in our paper in the following tables, including:

\begin{itemize}
    \item \textbf{Prompts for Synthetic Training Dataset:} The prompt is used by the training dataset constructed in \S \ref{sec:logic}. The same prompt format is adhered to when generating downstream domain data with our dataset.
    \item \textbf{Logic Generation Prompts For Extractive QA:} The prompt is used when generating Logic for extractive QA tasks.
    \item \textbf{Prompts for Synthetic Inspection Training Dataset:} The prompt is used when generating inspection data with DeepSeek-V3.
    \item \textbf{Different Task Prompts and Self-Inspection Prompts for AQuilt:} The prompt is used when generating downstream task data with AQuilt, covering all tasks we defined.
    \item \textbf{Prompts for General LLM Downstream Task Generation Based on Unlabeled Data:} The prompt is used when generating data based on unlabeled data with a general-purpose LLM.
    \item \textbf{Prompts for General LLM Downstream Task Generation using SelfInstuct:} The prompt is used when generating data with a general-purpose LLM using SelfInstuct.
    \item \textbf{Prompts for Evaluation on Downstream Tasks:} The prompt is used during the evaluation of downstream tasks.
    \item \textbf{GPT-4o Prompts for Independence Analysis in Synthetic Data:} The prompt is used when analyzing the independence of synthetic data with unlabeled data.
\end{itemize}

\begin{table*}
\begin{tcolorbox}[notitle, colback=gray!10,
colframe=black,
title={Prompts for Synthetic Training Dataset (Meta Prompts)},]
\texttt{
You are a professional Q\&A pair generation assistant. Your responsibility is to create complete, clear, accurate, and useful \{Task Type\} Q\&A pairs based on the provided text content. You need to deeply analyze the text to create reasonable questions and provide appropriate, detailed answers, ensuring that each Q\&A pair is relevant and useful.\\
I. The \{Task Type\} questions you create should meet the following requirements:\\
Requirement 1: \{Question Requirement\}\\
Requirement 2: The \{Task Type\} questions generated can be answered without <text>, and the question provides comprehensive information, complete context, including the core content or key information of <text>. Specifically: The question must explicitly contain the key information of <text> to ensure that the question itself is self-contained. - Do not rely on external context or assume that the user already understands the content of <text>. Avoid using phrases such as according to the above text, in the context, above content, or based on the information provided;\\
Requirement 3: The \{Task Type\} questions generated should be as complex as possible, requiring multi-step reasoning to determine the final answer;\\
Requirement 4: The generated answer is accurate and error-free, based on credible facts and data from the provided text;\\
Requirement 5: The generated answer is complete, not only selecting the correct option but also providing explanation and thinking steps to exclude other distractors;\\
Requirement 6: \{Thought Process Requirement\}\\
Requirement 7: When generating Q\&A pairs and thought process, assume there is no text as a reference, which means do not include phrases like the text, the context, or the information provided in the Q\&A pairs and thought process you create.\\
II. Please generate a Q\&A pair in the following format:\\
JSON\\
\{\\
question:\{\{The question you create\}\},\\
thought process:\{\{The thought process you create\}\},\\
answer:\{\{The answer you create\}\}\\
\}\\
\\
III. Please study the above requirements carefully and create a \{Task Type\} Q\&A pair:
}
\end{tcolorbox}
\end{table*}

\begin{table*}
\begin{tcolorbox}[notitle, colback=gray!10,
colframe=black,
title={Prompts for Synthetic Training Dataset (Specific Task Content 1/3)},]
\texttt{
\\
\textbf{Multi-Choice QA (single answer)}:\\
Task Type: single-choice\\
Question Requirement: The intent of the single-choice questions generated is clear and the semantics are explicit, including the question and necessary answers as well as distractors; \\
Thought Process Requirement: The thinking process should include the following steps: 1. Read the question: Understand the provided question. 2. Analyze the options: Assess the relationship and correctness of each candidate with the provided question. 3. Choose the best answer: Select the most accurate and contextually relevant option;\\
\\
\textbf{Multi-Choice QA (multi answer)}:\\
Task Type: multi-choice\\
Question Requirement: The intent of the generated multiple-choice question is clear and the semantics are explicit, including the question and necessary answers (there can be multiple answers that meet the requirements of the question) as well as distractors;\\
Thought Process Requirement: The thought process should include the following steps: 1. Read the question: Understand the question provided and clarify what is required. 2. Analyze the options: Assess each candidate option's relationship and correctness in relation to the provided question. 3. Choose reasonable answers: Based on the analysis of each option, select all reasonable answers;\\
\\
\textbf{Closed-Book QA}:\\
Task Type: closed-book\\
Question Requirement: The generated questions should be answerable without external knowledge and should provide comprehensive information, complete context, and contain relevant background information; do not generate questions about specific small events, as such questions are meaningless;\\
Thought Process Requirement: The thought process should include the following steps: 1. Understanding the question: Understand the question and clarify its requirements. 2. Analyzing the question: Analyze relevant information based on your own knowledge without relying on external resources. 3. Formulating a response: Construct a reasonable and accurate answer.\\
\\
\textbf{Open-Book QA}:\\
Task Type: open-book\\
Question Requirement: Open Q\&A refers to identifying and extracting specific information segments from the given text to answer the question. The generated question should explicitly include the text needed to answer the question and should not directly use the text provided by the user.\\
Thought Process Requirement: The generation of the thought chain should include the following steps: 1. Read the text: fully understand the problem and the provided text or paragraph. 2. Identify the relevant parts: locate the specific text segments that contain the answer to the question. 3. Construct the final reply: summarize the answer to the question.
}
\end{tcolorbox}
\end{table*}

\begin{table*}
\begin{tcolorbox}[notitle, colback=gray!10,
colframe=black,
title={Prompts for Synthetic Training Dataset (Specific Task Content 2/3)},]
\texttt{
\\
\textbf{Text Classification}:\\
Task Type: text classification\\
Question Requirement: The generated classification question should have a clear intent and unambiguous semantics, including the text content and predefined categories. \\
Thought Process Requirement: The generated thought process should include the following steps: 1. Analyze the content: Examine the content of the <text>, identifying themes, keywords, or other indicative features.2. Map to labels: Match the analyzed features to predefined labels or categories.3. Confirm classification: Verify that the assigned label accurately reflects the content.4. Record the result: Record or output the classification result.\\
\\
\textbf{Natural Language Inference}:\\
Task Type: natural language inference\\
Question Requirement:The question generated can be answered with Yes/No/Maybe.\\
Thought Process Requirement: The generated thought process should include the following steps:1.Understand the question: Understand the provided question. 2. Analyze the question: Identify which specific parts of the text the question is related to. 3. Logic Reasoning: provide logic reasoning to answer. 4. Provide the best answer: Select yes/no/maybe to answer the question.\\
\\
\textbf{Text Generation}:\\
Task Type: text generation\\
Question Requirement: The text generation problem should have a clear intent and be semantically clear. It must include the user's instructions and the conditions for generation.\\
Thought Process Requirement: The thought process for generation should include the following steps: 1. Understand the input conditions: Review the user's instructions and conditions to grasp the required scope, tone, and structure. 2. Brainstorm content: Develop key ideas or themes consistent with the input conditions. 3. Generate output: Create a well-structured and coherent response that follows the user's instructions.\\
\\
\textbf{Text Summarization}:\\
Task Type: text summarization\\
Question Requirement: The text summarization problem should have a clear intent and be semantically clear, including the text content and summarization requirements.\\
Thought Process Requirement: 1. Identify key points: Extract the most important information that represents the overall content of the source material. 2. Organize information: Logically arrange the extracted key points to ensure clarity and coherence. 3. Generate summary: Condense the key points into a concise and clear summary.
}
\end{tcolorbox}
\end{table*}

\begin{table*}
\begin{tcolorbox}[notitle, colback=gray!10,
colframe=black,
title={Prompts for Synthetic Training Dataset (Specific Task Content 3/3)},]
\texttt{
\\
\textbf{Natural Languaga Understanding}:\\
Task Type: natural language understanding\\
Question Requirement: The generated natural language understanding tasks can specifically include sentiment analysis, intent recognition, entity recognition, part-of-speech tagging, semantic analysis, etc. \\
Thought Process Requirement: The generated thinking steps should include the following steps: 1. Read the task: Understand the provided task. 2. Analyze the problem: Evaluate the relationship and correctness of the input text in relation to the task. 3. Provide the best answer: Respond with the most accurate and contextually relevant answer.\\
}
\end{tcolorbox}
\end{table*}

\begin{table*}
\begin{tcolorbox}[notitle, colback=gray!10,
colframe=black,
title={Logic Generation Prompts for Extractive QA},]
\texttt{
You are a professional assistant for generating thought process. Your responsibility is to synthesize useful thought process data based on the provided text content and question-answer pairs. You need to deeply analyze the text and construct a logical connection between the question and the answer.\\
1. The thought process you create should meet the following requirements:
Requirement 1: The generated thought process should be rich in content and logically clear, demonstrating how to infer the <Answer> from the <Question>.\\
Requirement 2: The thought process should include the following steps: (1). Read the question: Understand the provided question. (2). Analyze the question: Identify which specific parts of the text the question is related to. (3). Provide the best answer: Select the most relevant content from the text to answer the question. \\
2. Please generate the thought process in the following format:\\
JSON\\
\{\\
``thought\_process'': ``\{\{The thought process you created\}\}''\\
\}\\
3. Please carefully study the above requirements and then create a thought process based on the following <text>, <question>, and <answer> provided by the user:
}
\end{tcolorbox}
\end{table*}

\begin{table*}
\begin{tcolorbox}[notitle, colback=gray!10,
colframe=black,
title={Prompts for Synthetic Inspection Training Dataset (1/2)},]
\texttt{
You are an AI instruction and response quality assessment assistant, please score the quality of the user's instruction and response according to the following scoring criteria, and you can refer to the <text> provided by the user to evaluate the correctness of the question and answer pair:\\
1. The scoring criteria are as follows:\\
1 point - Low quality, minimal requirements met (Low Level):\\
- The response is only partially relevant to the question and lacks depth or detail.\\
- The response contains noticeable grammatical errors, spelling mistakes, or awkward phrasing.\\
- The response fails to address the user's questions or needs adequately.\\
- The response provides no additional explanation, context, or background information, leaving the user with limited understanding.\\
2 points - Basic requirements met (Qualified Level):\\
- The response is relevant and can basically meet the user's needs.\\
- The response has correct grammar and no obvious spelling errors.\\
- The response can solve the user's problem, but the solution may not be comprehensive or in-depth enough.\\
- The response provides basic explanations and background information, but not in detail.\\
3 points - Good quality, meeting most requirements (Good Level):\\
- The response is highly relevant and can well meet the user's needs.\\
- The response is fluent in grammar, without spelling errors, and clearly expressed.\\
- The response provides a comprehensive solution that can solve the user's problem and considers possible follow-up issues.\\
- The response provides detailed explanations and background information, which helps users understand.\\
4 points - High quality, meeting all requirements and exceeding expectations (Excellent Level):\\
- The response is highly relevant, not only meeting user needs but also anticipating and resolving potential issues.\\
- The response has perfect grammar, precise expression, and well-chosen words.\\
- The response provides an in-depth solution that can comprehensively solve the problem from multiple angles and provides additional useful information or suggestions.\\
- The response provides in-depth explanations and background information, which helps users gain a deeper understanding of the problem and solution.\\
5 points - Excellent quality, exceeding all requirements with professional contributions (Outstanding Level):\\
- The response is highly relevant, not only meeting user needs but also providing solutions beyond expectations.\\
- The response has impeccable grammar, elegant expression, and precise, impactful wording.\\
- The response provides an in-depth and professional solution that can solve the problem from a unique perspective and provides extremely valuable additional information or suggestions.
}
\end{tcolorbox}
\end{table*}

\begin{table*}
\begin{tcolorbox}[notitle, colback=gray!10,
colframe=black,
title={Prompts for Synthetic Inspection Training Dataset (2/2)},]
\texttt{
- The response provides in-depth explanations and background information, demonstrating a high level of professionalism and profound understanding of the issue.\\
2. Please analysis the quality in the following format:\\
JSON\\
\{\\
analysis\_steps: \{\{your analysis for the quality\}\},\\
score: \{\{your rate to the qa\_pair\}\}\\
\}\\
\\
3. Please carefully study the above scoring criteria and strictly follow the scoring criteria above to score the following <qa\_pair> based on the following <text> provided by the user:
}
\end{tcolorbox}
\end{table*}

\begin{table*}
\begin{tcolorbox}[notitle, colback=gray!10,
colframe=black,
title={Different Task Prompts for AQuilt (1/5)},]
\texttt{
\textbf{single choice question answering}: ``````Please generate a single-choice question from the provided reference materials to help students better grasp the relevant knowledge:
The single-choice question should include a question, four options labeled A, B, C, and D, one of which is the answer to the question;\\
At the same time, you also need to generate the thinking steps for solving the question, as well as the answer to this question.\\
And output in the following JSON format:\\
JSON\\
\{``question'': ``xxx'', ``thinking\_steps'': ``xxx'', ``answer'': ``xxx''\}\\
\\
""",\\
\textbf{multi choice question answering}: ``````Please generate a multiple-choice question from the references provided to help students better grasp the knowledge:\\
The multiple-choice question should include a question with multiple options tags A, B, C, D, E (and so on), one or more of which are the answers to the questions;
At the same time, you also need to generate the thinking steps for solving the question, as well as the answer to this question.\\
And output in the following JSON format:\\
JSON\\
\{``question'': ``xxx'', ``thinking\_steps'': ``xxx'', ``answer'': ``xxx''\}\\
"""\\
}
\end{tcolorbox}
\end{table*}

\begin{table*}
\begin{tcolorbox}[notitle, colback=gray!10,
colframe=black,
title={Different Task Prompts for AQuilt (2/5)},]
\texttt{
\\
\textbf{close-book question answering}: ``````Please generate a closed-book question and answer pair from the provided reference materials that do not require reference text to answer to help students better grasp the relevant knowledge:\\
This Q\&A pair should include a question, and you also need to generate the thinking steps for solving the question, as well as the answer to this question.\\
And output in the following JSON format:\\
JSON\\
\{``question'': ``xxx'', ``thinking\_steps'': ``xxx'', ``answer'': ``xxx''\}\\
""",\\
\textbf{open-book question answering}: ``````Please generate an open-book Q\&A pair from the provided reference materials to help students better grasp the relevant knowledge:
This Q\&A pair should include a question, and you also need to generate the thinking steps for solving the question, as well as the answer to this question.
And output in the following JSON format:\\
JSON\\
\{``question'': ``xxx'', ``thinking\_steps'': ``xxx'', ``answer'': ``xxx''\}\\
"""\\
}
\end{tcolorbox}
\end{table*}

\begin{table*}
\begin{tcolorbox}[notitle, colback=gray!10,
colframe=black,
title={Different Task Prompts for AQuilt (3/5)},]
\texttt{
\\
\textbf{text summarization}: ``````Please generate a concise summary Q\&A pairs of the provided text to help students better understand the main points:\\
The summary should capture the key ideas and essential information from the text.
The content you generate should include a question, and you also need to generate the thinking steps for solving the question, as well as the answer to this question.
And output in the following JSON format:\\
JSON\\
\{``question'': ``xxx'', ``thinking\_steps'': ``xxx'', ``answer'': ``xxx''\}\\
""",\\
\textbf{text generation}: ``````Please generate a text-generated Q\&A pair based on the text provided to help students learn:\\
The resulting text should be well-structured and relevant to the given text.
The content you generate should include a question, and you also need to generate the thinking steps for solving the question, as well as the answer to this question.
And output in the following JSON format:\\
JSON\\
\{``question'': ``xxx'', ``thinking\_steps'': ``xxx'', ``answer'': ``xxx''\}\\
"""\\
}
\end{tcolorbox}
\end{table*}

\begin{table*}
\begin{tcolorbox}[notitle, colback=gray!10,
colframe=black,
title={Different Task Prompts for AQuilt (4/5)},]
\texttt{
\\
\textbf{natural language inference}: ``````Please generate a logical inference question from the provided reference materials to help students better grasp the relevant knowledge:\\
Logical inference questions generally ask whether a judgment or piece of knowledge is correct, with answers including ``yes, no, maybe'' three options.\\
The content you generate should include a question, and you also need to generate the thinking steps for solving the question, as well as the answer to this question.\\
And output in the following JSON format:\\
JSON\\
\{``question'': ``xxx'', ``thinking\_steps'': ``xxx'', ``answer'': ``xxx''\}\\
""",\\
\textbf{text classification}: ``````Generate a text classification task based on the text provided to help students understand the content of the text:\\
Classifications should be accurate and relevant to the given text.\\
The content you generate should include a question, and you also need to generate the thinking steps for solving the question, as well as the answer to this question.\\
And output in the following JSON format:\\
JSON\\
\{``question'': ``xxx'', ``thinking\_steps'': ``xxx'', ``answer'': ``xxx''\}\\
"""\\
}
\end{tcolorbox}
\end{table*}

\begin{table*}
\begin{tcolorbox}[notitle, colback=gray!10,
colframe=black,
title={Different Task Prompts for AQuilt (5/5)},]
\texttt{
\\
\textbf{extractive question answering}: ``````Please generate an extractive question answering task based on the provided reference materials to help students better understand the main points:\\
The content you generate should include a question, and you also need to generate the thinking steps for solving the question, as well as the answer to this question.
And output in the following JSON format:\\
JSON\\
\{``question'': ``xxx'', ``thinking\_steps'': ``xxx'', ``answer'': ``xxx''\}\\
""",\\
\textbf{natural language understanding}: ``````Please generate a natural language understanding question (such as sentiment analysis, semantic analysis, entity recognition, etc.) based on the provided reference materials to help students better grasp the relevant knowledge:\\
The content you generate should include a question, and you also need to provide the thinking steps to solve the question, as well as the answer to the question.
Please output in the following JSON format:\\
JSON\\
\{``question'':``xxx'', ``thinking\_steps'': ``xxx'', ``answer'': ``xxx''\}\\
"""\\
}
\end{tcolorbox}
\end{table*}

\begin{table*}
\begin{tcolorbox}[notitle, colback=gray!10,
colframe=black,
title={Self-Inspection Prompts for AQuilt},]
\texttt{
Please score the quality of the user's instruction and response to help students understand the quality of the question and response based on the provided text.
There are 5 levels of quality, which are: 1 point, 2 points, 3 points, 4 points, 5 points. The higher the score, the better the quality.\\
You'll first need to analyze the quality of the question and response before grading it.
And output in the following JSON format:\\
JSON\\
\{``analysis\_steps'': ``xxx'', ``score'': ``xxx''\}\\
}
\end{tcolorbox}
\end{table*}

\begin{table*}
\begin{tcolorbox}[notitle, colback=gray!10,
colframe=black,
title={Prompts for General LLM Downstream Task Generation Based on Unlabeled Data (SquadQA)},]
\texttt{
You are a professional Q\&A pair generation assistant. Your responsibility is to create complete, clear, accurate, and useful extractive Q\&A pairs based on the provided text.\\
I. The Q\&A pairs you create should meet the following requirements:\\
Requirement 1: The generated questions should have clear intentions and semantics;\\
Requirement 2: The generated questions should be answerable without external knowledge; Do not rely on external context or assume that the user already understands the content of <text>.\\
Requirement 3: The question generated can be answered with the provided reference material.\\
Requirement 4: The generated answers should be accurate and based on credible facts and data;\\
Requirement 5: The generated answer can be found in the reference materials.\\
II. Please generate a Q\&A pair in the following format:\\
JSON\\
\{\\
``question'': ``\{\{The question you create\}\}'',\\
``answer'': ``\{\{The answer you create(Extracted from reference materials)\}\}''\\
\}\\
III. Please study the above requirements carefully and create a extractive Q\&A pair based on the <text> provided by the user below:\\
}\\
\end{tcolorbox}
\end{table*}

\begin{table*}
\begin{tcolorbox}[notitle, colback=gray!10,
colframe=black,
title={Prompts for General LLM Downstream Task Generation Based on Unlabeled Data (PubMedQA)},]
\texttt{
You are a professional Q\&A pair generation assistant. Your responsibility is to create complete, clear, accurate, and useful Yes/No Q\&A pairs based on the provided text.\\
I. The Q\&A pairs you create should meet the following requirements:\\
Requirement 1: The generated questions should have clear intentions and semantics;\\
Requirement 2: The generated questions should be answerable without external knowledge;\\
Requirement 3: The question generated can be answered with either Yes or No.\\
Requirement 4: The generated answers should be accurate and based on credible facts and data;\\
Requirement 5: The generated answer has only two options: Yes/No.\\
II. Please generate a Q\&A pair in the following format:\\
JSON\\
\{\\
``question'': ``\{\{The question you create\}\}'',\\
``answer'': ``\{\{The answer you create(Yes/No)\}\}''\\
\}\\
III. Please study the above requirements carefully and create a Yes/No Q\&A pair based on the <text> provided by the user below:\\
}
\end{tcolorbox}
\end{table*}

\begin{table*}
\begin{tcolorbox}[notitle, colback=gray!10,
colframe=black,
title={Prompts for General LLM Downstream Task Generation Based on Unlabeled Data (CEVAL)},]
\texttt{
You are a professional question-answer pair generation assistant. Your responsibility is to create complete, clear, accurate, and useful single-choice question-answer pairs based on the provided text content. You need to analyze the text in-depth, create reasonable questions, and provide appropriate and detailed answers to ensure that each question-answer pair is relevant and useful.\\
I. The single-choice questions you create should meet the following requirements:
Requirement 1: The single-choice questions should have clear intentions and be semantically clear, including the question and necessary options as well as distractors;\\
Requirement 2: The single-choice questions should be answerable without the <text> and the information provided in the question should be comprehensive, with complete context and relevant background information;\\
Requirement 3: The answers generated should be accurate and based on credible facts and data from the provided text;\\
Requirement 4: The answers generated should be complete and not omit any necessary information;\\
II. Please generate the question-answer pairs in the following format:\\
JSON\\
\{\\
``question'':``\{\{The question you create and the options\}\}'',\\
``answer'':``\{\{The correct answer you create\}\}''\\
\}\\
III. After carefully studying the above requirements, please create a single-choice question-answer pair based on the <text> provided below:
}
\end{tcolorbox}
\end{table*}

\begin{table*}
\begin{tcolorbox}[notitle, colback=gray!10,
colframe=black,
title={Prompts for General LLM Downstream Task Generation Based on Unlabeled Data (Translation)},]
\texttt{
You are an expert in generating question-and-answer pairs. Your task is to create complete, clear, accurate, and useful closed-book question-and-answer pairs based on the provided text content. You need to analyze the text in depth, formulate reasonable questions, and provide appropriate and detailed answers, ensuring that each question-and-answer pair is relevant and useful.\\
I. The question-and-answer pairs you create should meet the following requirements:\\
Requirement 1 The questions should have clear intentions and be semantically clear.\\
Requirement 2 The questions should be answerable without external knowledge, and the information provided in the question should be comprehensive and contextually complete.\\
Requirement 3 The questions should be legal translation questions. You need to first determine the language of the given text. If it is in Chinese, the question should be of the Chinese-to-English type. Conversely, if the given text is in English, the question should be of the English-to-Chinese type.\\
Requirement 4 The type of question you generate can be randomly selected from the following four options: ``Please translate the following sentence from the contract into Chinese/English:'', ``Please translate the following legal term into Chinese/English:'', ``Please translate the following sentence into Chinese/English:'', ``Please translate the following legal provision into Chinese/English:''\\
Requirement 5: The answers you generate should be accurate and based on reliable facts and data.\\
II. Please generate the question-and-answer pairs in the following format:\\
JSON\\
\{\\
``question'':``\{\{the question you create\}\}'',\\
``answer'':``\{\{the answer you create\}\}''\\
\}\\
III. After carefully studying the above requirements, please create a question-and-answer pair based on the <text> provided by the user below:\\
}
\end{tcolorbox}
\end{table*}

\begin{table*}
\begin{tcolorbox}[notitle, colback=gray!10,
colframe=black,
title={Prompts for General LLM Downstream Task Generation Based on Unlabeled Data (EassyQA)},]
\texttt{
You are a professional question-and-answer pair generation assistant. Your responsibility is to create complete, clear, accurate, and useful question-and-answer pairs based on the provided text content. You need to analyze the text in depth, create reasonable questions, and provide appropriate and detailed answers to ensure that each question-and-answer pair is relevant and useful.\\
I. The question-and-answer pairs you create should meet the following requirements:\\
Requirement 1: The generated questions should have clear intentions and be semantically clear.\\
Requirement 2: The questions should be answerable without external knowledge, and the information provided in the questions should be comprehensive and contextually complete.\\
Requirement 3: The questions should be legal essay questions, starting with: ``Please analyze the following essay question, elaborate on your views, and cite relevant legal provisions and principles. Ensure that you provide sufficient arguments and analysis for each question to clearly demonstrate your deep understanding and flexible application of legal issues.''\\
Requirement 4: The answers generated should be accurate and based on reliable facts and data.\\
II. Please generate the question-and-answer pairs in the following format:\\
JSON\\
\{\\
``question'': ``\{\{the question you create\}\}'',\\
``answer'': ``\{\{the answer you create\}\}''\\
\}\\
III. Please carefully study the above requirements and then create a Q\&A pair based on the <text> provided by the user.
}
\end{tcolorbox}
\end{table*}

\begin{table*}
\begin{tcolorbox}[notitle, colback=gray!10,
colframe=black,
title={Prompts for General LLM Downstream Task Generation using Self-Instruct (Input Generation w/o context)},]
\texttt{\\
As an InputGenerator , your task is to generate a new [input] based on the [instruction] and some example [input].\\
Try your best to ensure that the new [input] you generate is distinct from the provided [input] while maintaining a diverse, detailed, precise, comprehensive, and high-quality response. Avoid generating a new [input] that is the same as the provided [input].\\
Start of instruction\\
\{\{instruction\}\}\\
End of instruction\\
Here are some high-quality [input] for the [instruction]. These [input] can provide you with very strict format requirements .\\
Below are [N] [input] examples:\\
\{\{Input Examples\}\} \\
Please generate 1 [input] based on the examples and requirements:
}
\end{tcolorbox}
\end{table*}

\begin{table*}
\begin{tcolorbox}[notitle, colback=gray!10,
colframe=black,
title={Prompts for General LLM Downstream Task Generation using Self-Instruct (Input Generation w/ context)},]
\texttt{\\
As an InputGenerator , your task is to generate a new [input] based on the [instruction],  [context] and some example [input].\\
Try your best to ensure that the new [input] you generate is distinct from the provided [input] while maintaining a diverse, detailed, precise, comprehensive, and high-quality response. Avoid generating a new [input] that is the same as the provided [input].\\
Start of instruction\\
\{\{instruction\}\}\\
End of instruction\\
Here are some high-quality [input] and provided [context] for the [instruction]. These [input] can provide you with very strict format requirements .\\
Below are [N] [input] examples:\\
\{\{Input Examples\}\} \\
The provided context content is:
\{\{context\}\} \\
Please generate 1 [input] based on the examples, requirements, and the provided above context:
}
\end{tcolorbox}
\end{table*}

\begin{table*}
\begin{tcolorbox}[notitle, colback=gray!10,
colframe=black,
title={Prompts for General LLM Downstream Task Generation using Self-Instruct (Output Generation)},]
\texttt{\\
You are an AI question-answering bot, acting as an expert in the field of \{\{domain\}\}. Please refer to the question provided by the user and answer the question carefully.\\
User:Please answer the following question:\\
\{\{question\}\}
}
\end{tcolorbox}
\end{table*}

\begin{table*}
\begin{tcolorbox}[notitle, colback=gray!10,
colframe=black,
title={Prompts For evaluation on Downstream Tasks},]
\texttt{
\\
\textbf{SquadQA} \\
\\
Input: \{question\}\\
\\
\textbf{PubMedQA} \\
\\
Input: Context:\{context\} Based on the context above, please answer the following question:\{question\}\\
\\
\textbf{CEVAL} \\
\\
Input: Below is a multiple-choice question from a Chinese \{subject\} exam. Please select the correct answer.\\
\{question\}\\
A. \{A\} B. \{B\} C. \{C\} D. \{D\}\\
What is the answer?\\
\\
\textbf{Translation} \\
\\
Input: Please complete the following legal translation task and provide the translation directly. Translate the following \{text type\} into Chinese/English: \{legal text\}. \\
\\
\textbf{EssayQA}\\
\\
Input: Please analyze the following essay question. Elaborate on your views in detail and may cite legal provisions and relevant legal principles. Ensure that you provide sufficient arguments and analysis for each question to clearly demonstrate your profound understanding and flexible application ability of legal issues.\\
\{material\}\\
Question: \{question\}.
}
\end{tcolorbox}
\end{table*}

\begin{table*}
\begin{tcolorbox}[notitle, colback=gray!10,
colframe=black,
title={GPT-4o Prompts for Independence Analysis in Synthetic Data },]
\texttt{
You are a professional question analysis assistant, responsible for determining whether a question relies on a text for its answer based on the provided question.\\
Criteria for Judgment:\\
The question contains some obvious keywords that indicate reliance on a text, such as ``the above content,'' ``according to the text,'' ``the above text,'' ``in the text,'' ``in the passage,'' etc.\\
If the question is about understanding or inquiring about the content of a certain text, then it is also considered a question that relies on a text for its answer.\\
Formatting Requirements:\\
Please carefully review the above criteria.\\
Determine whether the question provided by the user has text dependency.\\
If it does, please answer directly with 'Yes.'\\
If it does not, please answer directly with 'No.
}
\end{tcolorbox}
\end{table*}

\end{document}